\documentclass{article}

\usepackage{arxiv}

\usepackage[utf8]{inputenc} 
\usepackage[T1]{fontenc}    
\usepackage{hyperref}       
\usepackage{url}            
\usepackage{booktabs}       
\usepackage{amsfonts}       
\usepackage{nicefrac}       
\usepackage{microtype}      
\usepackage{lipsum}		
\usepackage{graphicx}
\usepackage{natbib}
\usepackage{doi}
\usepackage{siunitx}

\title{Convolutional Xformers for Vision}

\author{ \href{https://orcid.org/0000-0003-4110-9638}{\includegraphics[scale=0.06]{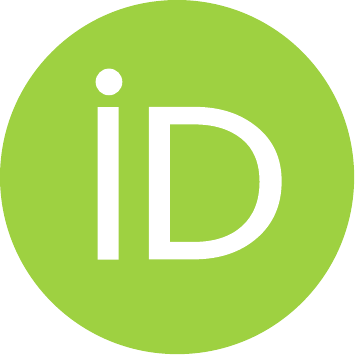}\hspace{1mm}Pranav Jeevan P} \\
	Department of Electrical Engineering\\
	Indian Institute of Technology Bombay\\
	Mumbai, India  \\
	\texttt{194070025@iitb.ac.in} \\
	\And
	\href{https://orcid.org/0000-0002-8003-6809}{\includegraphics[scale=0.06]{orcid.pdf}\hspace{1mm}Amit Sethi} \\
	Department of Electrical Engineering\\
	Indian Institute of Technology Bombay\\
	Mumbai, India \\
	\texttt{asethi@iitb.ac.in} \\
}

\date{}

\renewcommand{\headeright}{Preprint}
\renewcommand{\undertitle}{Preprint}
\renewcommand{\shorttitle}{Convolutional Xformers for Vision}

\hypersetup{
pdftitle={Convolutional Xformers for Vision},
pdfsubject={q-bio.NC, q-bio.QM},
pdfauthor={Pranav Jeevan, Amit Sethi},
pdfkeywords={Transformer, Image classification, Linear Attention, Deep Learning},
}

\begin{document}
\maketitle

\begin{abstract}
Vision transformers (ViTs) have found only limited practical use in processing images, in spite of their state-of-the-art accuracy on certain benchmarks. The reason for their limited use include their need for larger training datasets and more computational  resources compared to convolutional neural networks (CNNs), owing to the quadratic complexity of their self-attention mechanism. We propose a linear attention-convolution hybrid architecture -- Convolutional X-formers for Vision (CXV) -- to overcome these limitations. We replace the quadratic attention with linear attention mechanisms, such as Performer, Nyströmformer, and Linear Transformer, to reduce its GPU usage. Inductive prior for image data is provided by convolutional sub-layers, thereby eliminating the need for class token and positional embeddings used by the ViTs. We also propose a new training method where we use two different optimizers during different phases of training and show that it improves the top-1 image classification accuracy across different architectures. CXV outperforms other architectures, token mixers (e.g., ConvMixer, FNet and MLP Mixer), transformer models (e.g., ViT, CCT, CvT and hybrid Xformers), and ResNets for image classification in scenarios with limited data and GPU resources.  

\end{abstract}


\section{Introduction}
\label{sec:intro}
Even though transformers~\cite{vaswani2017attention,devlin2019bert} have become the state-of-the-art and at par with humans for several natural language processing (NLP) tasks, their applications in vision has been severely limited by their quadratic complexity with respect to sequence length. Even low resolution images, when unrolled, become long 1D sequences of tens of thousands of pixels, and impose a large computational and memory burden on a GPU. A transformer, being a general architecture without an inductive prior, also requires a large number of training images for giving good generalization compared to convolutional models. It also needs extra architectural changes, including the addition of positional embeddings, to gather the positional information of various image pixels. This demand for large amount of data and GPU resources is not suitable for resource-constrained scenarios where data and GPU capabilities are limited, such as green or edge computing \cite{khan2021transformers}. 

On the other hand, CNNs have the inductive priors, such as translational equivariance due to convolutional weight sharing and partial scale invariance due to pooling, to handle 2D images which enables them to learn from smaller datasets with less computational expenditure. But, they fail to capture long range dependencies compared to transformers and require deeper networks with several layers to increase their receptive fields. Combining the efficiency and inductive priors of CNNs with the long range information capturing ability of attention can create better architectures that are suitable for computer vision applications.

We propose a novel hybrid architecture, Convolutional Xformers for Vision (CXV)\footnote{Our Code is available at \href{https://github.com/pranavphoenix/CXV}{https://github.com/pranavphoenix/CXV}}  that uses both attention and convolutions for image classification. We use linear attention mechanisms in conjunction with convolutions in \emph{each layer} to capture both long range dependencies and utilise the inductive priors of convolutions. This combination uses less data and GPU. The convolutions provides the prior which eliminates the need to unroll the image into sequence of pixels and patches and adding positional  embeddings to keep the 2D positional information. The careful placement of convolutions and attention along with the layer normalizations enable our model to capture both spatial information and inter pixel relationships better. The resolution and size of the image is maintained throughout the layers. In other words, our model does not unroll the image before feeding to network nor does it need positional embeddings to capture location information.

Our architecture does not try to reduce the input sequence length using a hierarchical inverted pyramid architecture as done in other hybrid models, such as Convolutional vision Transformer (CvT)~\cite{wu2021cvt} and LeViT~\cite{graham2021levit}, nor does it use sequence-pooling as is done in Compact Convolutional transformers (CCT)~\cite{hassani2021escaping}. Instead we figured out an optimal arrangement of existing architectural elements, such as linear attention mechanisms, convolutions, layer normalizations, and residual connections to develop an efficient architecture than can give better generalization at reduced computational costs.

Additionally, we found that adding pre-norm to attention and feed-forward layers was harming the performance. Replacing it with a single normalization in each layer helped. We also introduce Dual Optimizer Training (DualOpT), a novel training strategy using two different optimizers at two different phases of training. Application of DualOpT in our experiments resulted in faster convergence and higher accuracy compared to using a single optimizer. Our extensive ablation studies provide several new insights about designing novel transformers for images.

We compare our model with most of the established architectures for image classification, including convolutional models (ResNet-18 and ResNet-34~\cite{he2015deep}), token mixing models  (FNet~\cite{leethorp2021fnet}, MLP-Mixer~\cite{tolstikhin2021mlpmixer}, WaveMix~\cite{p2022wavemix} and ConvMixer~\cite{anonymous2022patches}), and vision transformers (ViT~\cite{dosovitskiy2021image}, Hybrid Vision Nyströmformer (ViN), Hybrid Vision Performer (ViP)~\cite{jeevan2021vision}, Hybrid Vision Linear Transformer (ViLT), Compact Convolutional Transformer (CCT)~\cite{hassani2021escaping} and Convolutional vision Transformer (CvT)~\cite{wu2021cvt}). Our experiments show that CXV outperforms all of them in low resource settings for the same number of parameters. When we compare the GPU usage and accuracy, we find that our CXV provides improved accuracy with less data without consuming as much GPU memory and computations compared to the other models.  We created a new hybrid Xformer architecture, Hybrid Vision Linear Transformer (Hybrid ViLT) using Linear Transformer~\cite{katharopoulos2020transformers} as the linear attention mechanism. Hybrid ViLT performs on par with other hybrid xformers (hybrid ViN and hybrid ViP) but consumes only half of the GPU compared to them.

\section{Related Works}

ViT~\cite{dosovitskiy2021image} overcame the quadratic complexity of self-attention by breaking the image into patches of size 16x16, which significantly reduced the sequence length upon unrolling. It also used positional embedding to capture the spatial  information of patches and added a learnable class embedding to capture the global image representation for classification. But, ViT still suffered from requirement for a large amount of data to provide good generalization. Additionally, ViT needed GPUs with large memories to fit state-of-the-art models, which also prevented its widespread use. 

CCT~\cite{hassani2021escaping}, LeViT~\cite{graham2021levit} and CvT~\cite{wu2021cvt} used convolutions to provide inductive priors to the vision transformers, which was shown to improve their performance. LeViT replaced patch-wise projections with convolutional embeddings and used 2D relative positional biases instead of initial absolute positional bias used in ViT. It also down-sampled the image in stages using an extra non-linearity in attention and also replaced layer-norm with batch-norm. CvT used a hierarchy of transformers containing a new convolutional token embedding, and a convolutional transformer block. CCT eliminated the class token and positional embeddings and used a sequence pooling strategy and convolutions for better performance. CCT also showed that transformer models can be trained to achieve state-of-the-art generalization using less data. Hybrid Xformers~\cite{jeevan2021vision} replaced the quadratic attention with linear attention mechanisms such as Performer~\cite{choromanski2021rethinking} and Nyströmformer~\cite{xiong2021nystromformer}, and also used convolutional layers to generate embeddings. They showed good performance with significant reduction in GPU usage. The Hybrid Vision Linformer (ViL) was shown to perform poorly in their experiments on image classification. Therefore, we used the hybrid xformer architecture to create Hybrid Vision Linear Transformer (Hybrid ViLT), where we used Linear Transformer as a linear attention module. 

Search for alternative architectures that mix information between tokens like self-attention, but require fewer computations and GPU memories, and require less data to generalize has led to the development of many token-mixer models. FNet~\cite{leethorp2021fnet} replaces the self-attention layer in the transformer with the Fourier transform and has shown to perform well compared to self-attention models, such as BERT~\cite{devlin2019bert}. We use FNet for image classification similar to experimentation in ~\cite{jeevan2021vision}. MLP-mixer~\cite{tolstikhin2021mlpmixer} replaces self-attention using two multi-layer perceptrons (MLPs) which are applied independently, first to image patches and then across patches. ConvMixer~\cite{anonymous2022patches} works similar to MLP-Mixer but uses standard convolutions instead of MLPs for mixing of information in spatial and channel dimensions. We compare the performance and GPU usage of our CXV models with these different architectures and show that CXV performs better in low-data low-GPU settings. 

A jump in accuracy was obtained by convolutional and recurrent networks while training when optimizer was switched from Adam to Stochastic gradient descent (SWATS) for various machine learning tasks~\cite{keskar2017improving}.

\section{CXV Model Architecture}

\begin{figure}[h]
\centering
\includegraphics[scale=0.85]{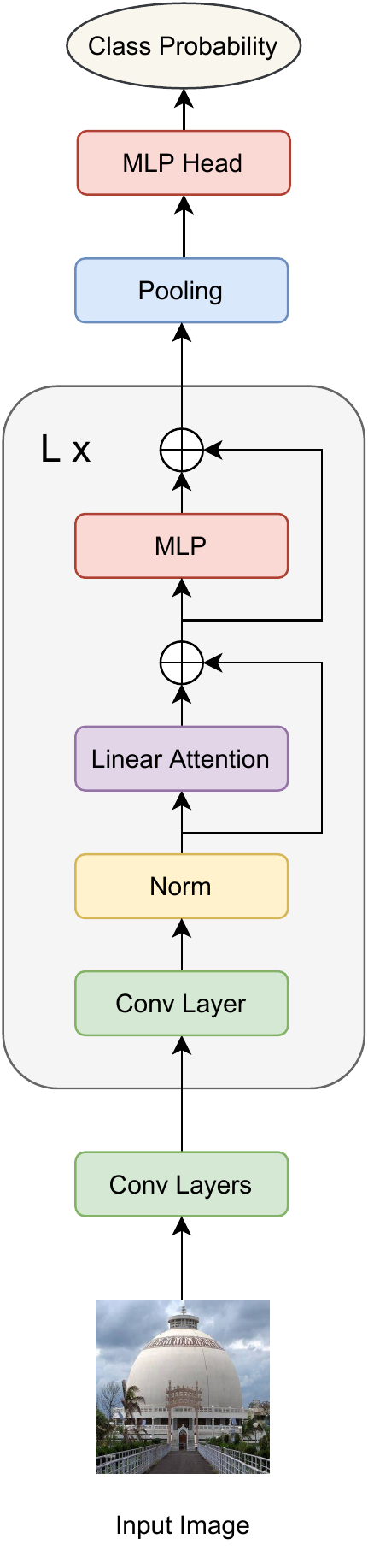}
\caption{Architecture of CXV}
\label{fig:CXV}
\end{figure}

The overall architecture of Convolutional Xformer for Vision (CXV) is shown in Fig. \ref{fig:CXV}. Before describing the overall architecture in detail, we explain the motivation behind and the design of some of its key components.



\subsection{Linear Attention}

It has been shown that replacing quadratic attention in ViT with linear attention mechanisms~\cite{jeevan2021vision}, such as Nyströmformer~\cite{xiong2021nystromformer} and Performer~\cite{choromanski2021rethinking} can reduce their GPU consumption considerably while improving their performance in low data image classification. The use of Xformers will enable us to handle longer image pixel sequences without incurring much computational costs. We use Nyströmformer, Performer, and Linear Transformer~\cite{katharopoulos2020transformers} in our experiments.

\subsection{Convolutional Embedding}
It has been shown that using convolutional layers to generate the pixel embeddings is better~\cite{jeevan2021vision,graham2021levit} than using the linear layers as used in ViT~\cite{dosovitskiy2021image} as the convolutional layers provide the inductive prior suitable for images to the transformer network significantly reducing the need for more training data. We also use convolutional layers to generate the embeddings which will be fed to the subsequent layers of the network. 
In case of high resolution images, kernel size and stride can be adjusted in these convolutional layers so that the sequence length of image pixel remains within the range as fixed by the available GPU size.

\subsection{Convolutional Sub-layer}
With Convolutional vision Transformer (CvT) architecture \cite{wu2021cvt}, it was shown that adding a convolutional sub-layer in the transformer layer improves performance and robustness, and it also maintains a high degree of computational and memory efficiency. But, CvT is a multi-stage hierarchical architecture where convolutional token embedding performed an overlapping convolution operation with stride on a 2D-reshaped
token map. It also progressively decreased the sequence length while simultaneously increasing the dimension of token features across stages. They also used one convolutional sub-layer per stage (three overall) instead of adding one per layer.

We use a convolutional sub-layer in our transformer layers to provide the spatial information and inductive priors to the linear attention sub-layer that follows. The use of convolutional sub-layer enables the architecture to maintain the 2D image shape throughout the network, except within the linear attention sub-layer where it is unrolled as a sequence of image tokens. The presence of convolutional sub-layer also eliminates the need for additional positional embeddings which are otherwise needed in other ViT architectures.

\subsection{Layer Normalization}

Even though the original transformer~\cite{vaswani2017attention} used post-norm, where layer normalization occurs after both attention and feed-forward sub-layers in order to reduce the variance of the inputs to the following sub-layer, Vision Transformers~\cite{dosovitskiy2021image} have been using pre-norm residual units, where layer normalization occurs before attention and feed-forward sub-layers. Post-norm transformers were observed to have larger magnitude gradient in later layers compared to initial layers~\cite{xiong2020layer}. Pre-norm has been shown to make back-propagation more efficient in training deep transformer models and often yielded improved performance.

It has been shown that in shallow transformer architectures ($\leq 6$ layers), pre-norm tends to degrade the performance \cite{Nguyen2019TransformersWT}. LeViT architecture removes the pre-norm completely from the attention and feed-forward sub-layers for faster inference and only uses a batch normalization in the convolutional layers~\cite{graham2021levit}. \cite{shleifer2021normformer} showed that while pre-norm improves stability over post norm, it creates the opposite issue on gradients; gradients at earlier layers tend to be larger than gradients at later layers.

In our CXV architecture, we remove pre-norm used in ViT and replace it with a single layer normalization placed between the convolutional block and self-attention in the each layer. The presence of one layer normalization per layer instead of two is enough for shallow networks. 

\subsection{Overall Architecture of CXV}

As shown in Fig.~\ref{fig:CXV}, the input image is fed to the convolutional layers, which generates the embeddings of an appropriate dimension. We use two convolutional layers to increase the channel dimension from three to the required embedding dimension. The kernel and strides are chosen to ensure that 32$\times$32 sized 2D feature maps are generated and the number of feature maps matches the embedding dimension.

The output from the convolutional embedding layer is then send to the convolutional xformer layers. In each layer, the input is passed through a convolutional sub-layer of kernel size 3x3, with stride and padding of 1. Then its output is normalized in the channel dimension in the normalization sub-layer. The output after normalization is given to the linear attention sub-layer where it is processed by Performer, Nyströmformer or Linear Transformer. The output from attention sub-layer is finally passed through an multi-layer perceptron (MLP) before it is send to the next layer. Residual connections~\cite{he2015deep} are used with both the linear attention and MLP sub-layers.

The output from the final layer is passed through a pooling layer, where it is average pooled and fed to an MLP head, which gives the output class. The resolution and size of input remains the same throughout the CXV layers. The 2D structure of the image is also maintained everywhere except within the linear attention mechanism.

\subsection{DualOpT: Dual-Optimizer Training}
We implemented a new training paradigm, Dual-Optimizer training (DualOpT) where two different optimizers where used to train the model at different phases of training. During the initial epochs of training, we used AdamW optimizer which showed faster convergence to final validation accuracy. Once the top-1 accuracy converged to a fixed value which remained unchanged for 20 epochs, we changed the optimizer to SGD. This change showed a sudden improvement in accuracy within a few epochs. The number of epochs needed in the training phase of the second optimizer is much shorter than the first optimizer for reaching the final top-1 accuracy. The accuracy obtained was much higher than the value obtained by using either of the two optimizers alone.

\section{Experiments and Results}

Our experiments were conducted using multiple data sets, multiple comparative neural architectures, and variations of our own architecture for ablation studies.

\subsection{Datasets and Other Models Compared}

The CIFAR-10 (MIT License), CIFAR-100 (MIT License)~\cite{Krizhevsky09learningmultiple}, and the Tiny ImageNet (MIT License)~\cite{Le2015TinyIV} datasets were used in our experiments to compare the performance of different models in low resource scenario. We chose ResNet-18 and ResNet-34~\cite{he2015deep} from pure convolutional models; FNet~\cite{leethorp2021fnet}, MLP-Mixer~\cite{tolstikhin2021mlpmixer}, WaveMix~\cite{p2022wavemix}, and ConvMixer~\cite{anonymous2022patches} from token mixing models; and ViT~\cite{dosovitskiy2021image}, Hybrid ViN, Hybrid ViP~\cite{jeevan2021vision}, Hybrid ViLT, CCT~\cite{hassani2021escaping}, and CvT~\cite{wu2021cvt}\footnote{Base code: \href{https://github.com/lucidrains/vit-pytorch}{https:github.com/lucidrains/vit-pytorch}} from transformer models for comparison with our CXV. Except for ResNets and MLP Mixer, all other models with similar number of parameters ($\sim1.3 $ M) as the CXV were used in our experiments. The number of layers, heads and embedding dimensions were adjusted to keep the numbers of parameters same. 

For the model nomenclature, we use the name of the model followed by the number of layers and number of heads as \emph{Model-layer/heads}, e.g., the CCT model with 6 layers and 8 heads is CCT-6/8. For models that does not have multiple heads, we omit the heads part in the model name, e.g., FNet having 8 layers is FNet-8. For CXV model names, we replace the X (Xformer) with the linear attention mechanism used in the model, e.g., CPV is Convolutional Performer for Vision.

Except ConvMixer, MLP-Mixer, and WaveMix, all other models used in our experiments have 128-dimensional embeddings. ConvMixer and MLP-Mixer used 256-dimensional embeddings and WaveMix used 64 dimension embeddings. A dropout of 0.5 was used across all models. We used 64 benchmark points in Nyströmformer and a local window size of 256 with ReLU non-linearity as the kernel function in Performer. 

For the DualOpT training of the proposed CXV models, AdamW optimizer with $\alpha = 0.001$ (learning rate), $\beta_{1} = 0.9$ and $\beta_{2}=0.999$ were used for computing running averages of gradient and its square, $\epsilon = 10^{-8}$, and 0.01 as weight decay coefficient was used in the initial phase of training and SGD (stochastic gradient descent) with learning rate of $0.001$ and momentum $= 0.9$ was used for the final phase. 

We used automatic mixed precision in PyTorch to make the training faster and consume less GPU memory. Experiments were done with 16 GB Tesla P100-PCIe and Tesla T4 GPUs available in Kaggle and Google Colab. Number of model parameters,  MACs, top-1 \% accuracy and GPU usage for a batch size of 32 are reported. All reported top-1 \% accuracy values are best out of three runs based on accepted reporting protocols~\cite{hassani2021escaping}. 

All models use MLP dimension that was double the size of its input embedding dimension. Only WaveMix architecture used 32 as MLP dimension. The hybrid xformers used three convolutional layers with 32, 64, and 128 kernels respectively of size 3$\times$3 with stride of one with padding to generate the embeddings. We used learnable position embedding and 3$\times$3 convolutional and pooling kernels having stride equal to 1 and padding in CCT. For CvT, we used two attention modules in the first and second stages and one in the third stage with 3$\times$3 convolutions in each stage without down-sampling.

While training with Tiny ImageNet dataset, the stride of initial convolution layer was adjusted to create model of same size as those used for CIFAR-10 and CIFAR-100. In ViT, we used a patch size of 2$\times$2 for training Tiny ImageNet. All models were trained on the three datasets without using any data augmentations or warm up. RandAugment~\cite{cubuk2019randaugment} was used only for the CXV later to see the improvements from augmentations (see ablation studies).

\subsection{Image Classification Results}

\label{subsec:img}

\begin{table*}[h]
\centering
\begin{tabular}{lcccccc}
\toprule
\multicolumn{1}{l|}{\textbf{Model}} &
  \textbf{\# Params} &
  \textbf{GPU (GB)} &
  \multicolumn{1}{c|}{\textbf{MACs}} &
  \textbf{CIFAR-10} &
  \textbf{CIFAR-100} &
  \textbf{Tiny ImageNet} \\ \midrule
\multicolumn{7}{l}{\textit{Convolutional Models}}                                                                                    \\ \midrule
\multicolumn{1}{l|}{ResNet-18}       & 11.2 M & 0.6          & \multicolumn{1}{c|}{0.56 G} & 86.29 & 59.15          & 43.02          \\
\multicolumn{1}{l|}{ResNet-34}       & 21.3 M & 0.7          & \multicolumn{1}{c|}{1.16 G} & 87.97 & 56.05          & 42.65          \\ \midrule
\multicolumn{7}{l}{\textit{Mixing Models}}                                                                                           \\ \midrule
\multicolumn{1}{l|}{FNet-10}            & 1.3 M  & 2.3          & \multicolumn{1}{c|}{1.34 G} & 51.05 & 11.97          &   8.14             \\
\multicolumn{1}{l|}{MLP Mixer-5}       & 21.3 M & 1.5          & \multicolumn{1}{c|}{3.02 G} & 60.26 & 34.81          & 20.26          \\
\multicolumn{1}{l|}{WaveMix-5}         & 1.4 M  & 0.4 & \multicolumn{1}{c|}{2.59 G} & 83.71 & 51.62          &    32.70          \\
\multicolumn{1}{l|}{ConvMixer-16}       & 1.3 M  & 3.5          & \multicolumn{1}{c|}{1.37 G} & 88.46 & \textbf{61.80} & 45.39          \\ \midrule
\multicolumn{7}{l}{\textit{Transformer Models}}                                                                                      \\ \midrule
\multicolumn{1}{l|}{ViT-10/4}        & 1.3 M  & 14.7         & \multicolumn{1}{c|}{1.34 G} & 57.53 & 30.80           & 23.18          \\
\multicolumn{1}{l|}{Hybrid ViN-6/8}  & 1.3 M  & 5.3          & \multicolumn{1}{c|}{1.41 G} & 77.96 & 50.06          & 36.84          \\
\multicolumn{1}{l|}{Hybrid ViP-6/8}  & 1.3 M  & 5.9          & \multicolumn{1}{c|}{1.31 G} & 77.54 & 57.22          & 38.85               \\
\multicolumn{1}{l|}{Hybrid ViLT-6/8} & 1.3 M  & 2.6          & \multicolumn{1}{c|}{0.33 G} & 78.34 & 52.38          & 37.17          \\
\multicolumn{1}{l|}{CCT-6/4}         & 1.3 M  & 13.6         & \multicolumn{1}{c|}{1.32 G} & 82.66 & 58.34          & 35.21 \\
\multicolumn{1}{l|}{CvT-5/4}         & 1.3 M  & 9.4          & \multicolumn{1}{c|}{1.32 G} & 77.90 & 57.82          & 39.81          \\ \midrule
\multicolumn{7}{l}{\textit{\textbf{Convolutional Xformers for Vision}}}                                                              \\ \midrule
\multicolumn{1}{l|}{CNV-5/4} &
  1.3 M &
  3.1 &
  \multicolumn{1}{c|}{1.39 G} &
  89.56 &
  59.23 &
 \textbf{49.56} \\
\multicolumn{1}{l|}{CLTV-5/4} &
  1.3 M &
  1.8 &
  \multicolumn{1}{l|}{1.38 G} &
  86.99 &
  60.11 &
  46.69 \\
\multicolumn{1}{l|}{CPV-5/4} &
  1.3 M &
  3.2 &
  \multicolumn{1}{c|}{1.37 G} &
  \textbf{91.42} &
  57.34 &
  47.29 \\ \bottomrule
\end{tabular}

\caption{Comparison of top-1 accuracy and computational costs for all the models in image classification using different datasets. The GPU consumption shown is for a batch size of 32. Warm up and data augmentations were not used for training.}
\label{tab:results}
\end{table*}

Table~\ref{tab:results} shows the top-1 accuracy of all the models trained on the each of the three datasets. The results shows that CXV models generalize better while also consuming lesser GPU memory compared to all the other models for the  three datasets. CXV models require orders of magnitude fewer parameters compared to ResNets and yet they outperform ResNets in image classification by 6\% in CIFAR-10 and 16\% in Tiny ImageNet datasets. 

\begin{figure}[h]
\centering
\includegraphics[scale=0.85]{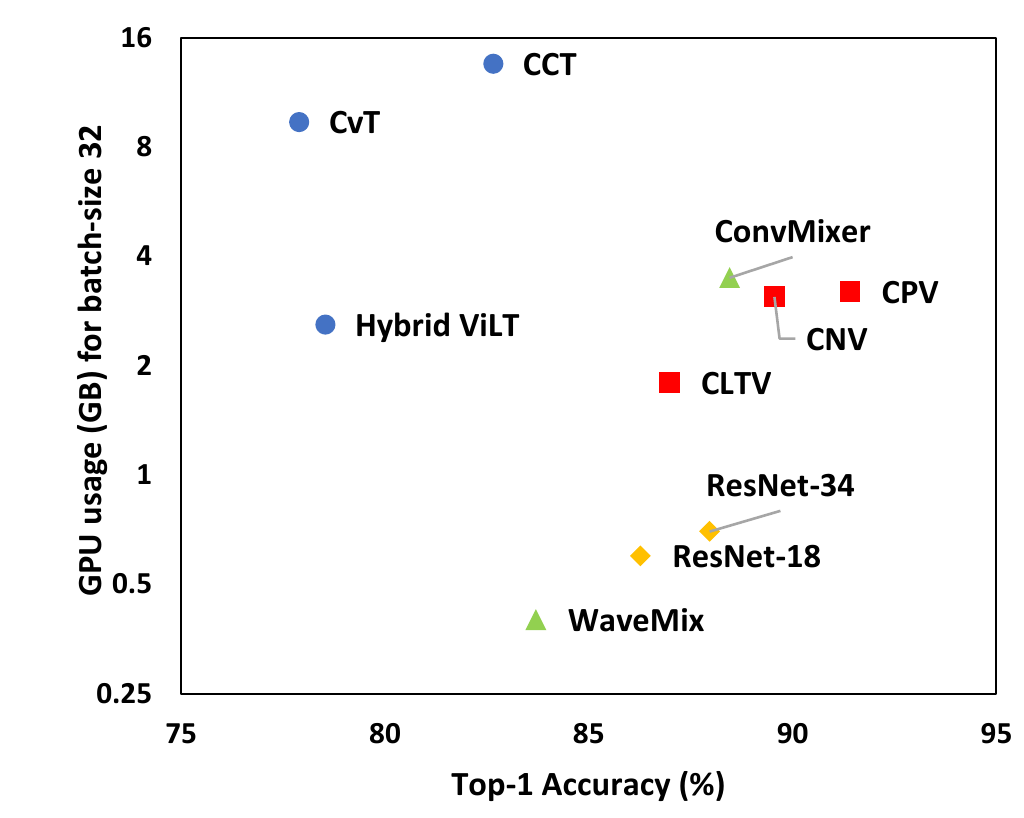}
\caption{Comparison of Top-1 Accuracy with GPU usage for various models on CIFAR-10 dataset}
\label{fig:GPU}
\end{figure}

Fig. \ref{fig:GPU} shows that token-mixing models (green triangles) consume very less GPU memory compared to transformer models (blue circles) and they still perform comparable to self-attention. WaveMix and ConvMixer outperformed conventional transformer architectures by significant margin. ConvMixer performed better than CXVs in CIFAR-100 dataset but it consumed almost double the GPU RAM consumed by CLTV (Convolutional Linear Transformer for Vision). WaveMix was the most GPU efficient model among all models tested. Both FNet and MLP-Mixer performed poorly compared to other models even though they consumed very little GPU. Even though MLP-Mixer had similar number of parameters compared to ResNet-34, it consumed twice the GPU memory and under-performed compared to it, which shows, once again, that CNNs are better at handling images than MLPs.

Among the transformer models, we observe that hybrid ViX models consume the least amount of GPU RAM while giving comparable performance to CCT and CvT. CXV models (red squares) perform $\approx 10\%$ better in CIFAR-10 while consuming two orders of magnitude less GPU RAM.

\section{Ablation Studies}

CXV models were tested extensively using variety of architectural modifications to build the optimal architecture.

\subsection{Architectural Variations}

\begin{figure}[h]
\centering
\includegraphics[scale=0.85]{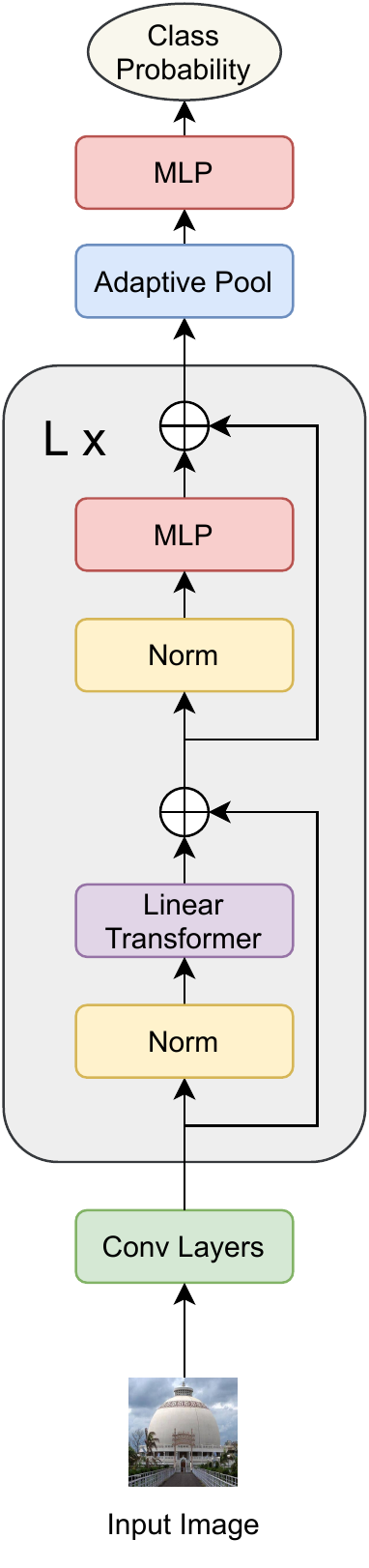}
\caption{Architecture of Hybrid ViLT}
\label{fig:ViLT}
\end{figure}

We compare the performance of CXV with other efficient transformers, such as hybrid vision nystromformer and hybrid vision performer that were shown to perform well, but hybrid vision linformer performed poorly in experiments. So we replaced hybrid vision linformer with another linear attention mechanism, the linear transformer. We used the hybrid xformer architecture used by~\cite{jeevan2021vision} and used Linear Transformer~\cite{katharopoulos2020transformers} as the attention mechanism. Linear transformer express self-attention as a linear dot-product of kernel feature maps and uses the associative property of matrix products to reduce complexity. To suit the image data, Q, K and V matrices were created using convolutional operations\footnote{Base code: \href{https://github.com/lucidrains/linear-attention-transformer/blob/master/linear_attention_transformer/images.py}{https://github.com/lucidrains/linear-attention-transformer}} This allowed us to pass the input in 2D form to the attention sub-layer eliminating the need for the extra class token and positional embeddings. The architecture of Hybrid ViLT (Vision Linear Transformer) is shown in Fig.~\ref{fig:ViLT}. Hybrid ViLT model consumed less than half the GPU used by other hybrid ViX models and performed almost comparable to them in all three datasets, as shown in Table~\ref{tab:results} and Fig.~\ref{fig:GPU}. 

Reducing the resolution of the input image by removing padding from initial convolution embedding layers showed around 17\% reduction in GPU usage but only a reduction in accuracy by less than 2\%.

Table~\ref{tab:cnn} shows that accuracy first increases and then starts to decrease as we increase the number of convolutional layers for creating the embeddings.
\begin{table}[]
\centering
\begin{tabular}{@{}lll@{}}
\toprule
1 Layer & 2 Layer & 3 Layer \\ \midrule
89.45   & 89.93   & 89.56   \\ \bottomrule
\end{tabular}
\vspace{2mm}
\caption{Variation of accuracy with number of initial convolutional embedding layers in CNV for classification of CIFAR-10 dataset.}
\label{tab:cnn}
\end{table}

We experimented with using GeLU activation in between the initial convolutional embedding layers and observed a decrease in accuracy by about 2 percentage points. When we used GeLU after the convolution sub-layer in each layer, it was also shown to decrease the final accuracy of the model.  A similar decrease in accuracy was observed when we used 2D max-pooling as used in CCT model.  

We also experimented with decreasing the convolutional sub-layers by using only one convolutional sub-layer for every two layers and observed a degradation in performance. So we concluded that placement of a convolutional sub-layer is essential for improvement in performance.

We also tried different combinations of residual connections by using residual connection around the convolutional sub-layer and around both the convolutional and layer normalization sub-layers. These extra residual connection did not lead to increase in accuracy. This may be due to the shallowness ($\leq 6$ layers) of the networks and these connections might become essential while using deeper networks with more layers. 

Since the network processes the images mostly in 2D and unrolling happens only within the attention sub-layers, it was observed that the accuracy did not improve by addition of positional embeddings (both 1D learnable~\cite{dosovitskiy2021image} and rotary~\cite{su2021roformer}). This might be due to the presence of the convolutional sub-layers which possess the 2D inductive priors needed to handle 2D images and they provide the information to the attention sub-layers eliminating the need for extra positional information. 

Replacing layer-normalization with batch-normalization was shown to reduce the accuracy by over four percentage points. We tried to reduce the number of layer-normalization by placing a layer-norm sub-layer in every alternate layers and observed the loss does not converge when there is an absence of normalization in each layer. Adding pre-norm to attention and feed-forward sub-layers was found to degrade the performance considerably.

\subsubsection{DualOpT}
\begin{table}[]
\centering
\begin{tabular}{@{}lccc@{}}
\toprule
\multicolumn{1}{c}{Models} &
  \begin{tabular}[c]{@{}c@{}}Accu. with \\ AdamW\end{tabular} &
  \begin{tabular}[c]{@{}c@{}}Accu. with \\ DualOpT\end{tabular} &
  \begin{tabular}[c]{@{}c@{}}\% \\ Increase\end{tabular} \\ \midrule
\multicolumn{4}{l}{Tiny ImageNet}   \\ \midrule
ResNet-34   & 39.93 & 42.65 & 6.81  \\
Hybrid ViN  & 34.84 & 36.84 & 5.74  \\
Hybrid ViP  & 35.20 & 38.85 & 10.37 \\
ViT         & 20.86 & 23.18 & 11.12 \\
WaveMix     & 31.81 & 33.04 & 3.87  \\
CLTV        & 44.49 & 46.69 & 4.94  \\ \midrule
\multicolumn{4}{l}{CIFAR-100}       \\ \midrule
Hybrid ViN  & 48.61 & 50.06 & 2.98  \\
Hybrid ViLT & 49.75 & 52.38 & 5.29  \\
CLTV        & 57.62 & 60.11 & 4.32  \\
CvT         & 54.24 & 57.82 & 6.60  \\ \midrule
\multicolumn{4}{l}{CIFAR-10}        \\ \midrule
Hybrid ViLT & 76.42 & 78.34 & 2.51  \\
CPV         & 90.55 & 91.42 & 0.96  \\
CLVT        & 85.10 & 86.82 & 2.02  \\ \bottomrule
\end{tabular}
\vspace{2mm}
\caption{Percentage increase in the top-1 accuracy of few models while using DualOpT compared to accuracy when only one optimizer is used across different datasets.}
\label{tab:DualOpT}
\end{table}

Table~\ref{tab:DualOpT} shows the comparison of top-1 accuracy obtained by few models while using one optimizer (AdamW) and the performance gain obtained while using DualOpT (AdamW + SGD). We observed similar performance gains across all models trained in all the three datasets. The accuracy obtained by DualOpT was higher than what was obtained when we trained using only AdamW and only SGD. AdamW was shown to cause convergence during initial phase of training and SGD was shown to be faster towards the final phase. While implementing DualOpT, it was observed across models that when shifting from AdamW to SGD, the training and testing accuracy jumps by a significant margin in the first epoch and the speed of convergence increases significantly. The improvement in performance using DualOpT was also observed while using RandAugment.

\subsection{Data Augmentation}

\begin{table}[]
\centering
\begin{tabular}{@{}ccc@{}}
\toprule
Model & \begin{tabular}[c]{@{}c@{}}Accuracy with \\ RandAugment (\%)\end{tabular} & \begin{tabular}[c]{@{}c@{}}Accuray after \\ Post-training (\%)\end{tabular} \\ \midrule
CNV-5/4  & 91.98 & 92.26 \\
CPV-5/4  & 92.21 & \textbf{94.46} \\
CLTV-5/4 & 88.16 & 91.40 \\ \bottomrule
\end{tabular}
\vspace{2mm}
\caption{Performance of CXV models using RandAugment and after further post-training using the un-augmented CIFAR-10 dataset.}
\label{tab:rand}
\end{table}

We assessed the effect of data augmentation using RandAugment~\cite{cubuk2019randaugment} on CXV models with number of augmentation ($M$) and magnitude of augmentations ($N$) set to one for CNV and CPV, and three for CLTV. The model parameters were same as used in Section~\ref{subsec:img}. The accuracy increased significantly when RandAugment was used as shown in Table~\ref{tab:rand}.

It was observed during training that training accuracy was always stagnating at a lower value compared to its corresponding value when training without using RandAugment, even after test accuracy converged to its final value. So, after the final test accuracy was achieved, we trained the model using the dataset without RandAugment. This post-training on the un-augmented dataset caused a further increase in performance of the model as shown in Table~\ref{tab:rand}. Both training using RandAugment and post-training without RandAgument was done using DualOpT. 

We conjecture that the increase in accuracy observed in post-training to be the fine-tuning effect created from the model seeing the un-augmented data for the first time which is equivalent to seeing a new dataset. This procedure of training improved the accuracy of CXV models by $\approx 3$ percentage points. We believe that the results reported can be further improved with more extensive tuning of the hyper-parameters. 

\section{Conclusions and Discussion}
We introduced a new hybrid neural architecture for processing images that combines the inductive priors from convolutions and self-attention from transformers that requires smaller training data and less GPU memory for state-of-the-art generalization. Our results using linear transformers for image classification agrees with previous studies~\cite{jeevan2021vision} that using linear attention mechanism can reduce the GPU consumption without affecting generalization. Adding convolutional sub-layers in our architecture ensured that our CXV models out-perform the current data-efficient architectures, such as CCT and CvT, considerably using one fourth the GPU memory for a given batch size. For our experiments we also constructed a new hybrid ViX model, the hybrid Vision Linear Transformer (ViLT), and showed that it consumes even lesser GPU RAM than other hybrid ViX models mentioned in~\cite{jeevan2021vision}. Our CXV models use order of magnitude fewer parameters than convolutional models, such as ResNets while also requiring orders of magnitude lesser GPU RAM than transformer models. Our CXV models can be used for training computer vision models in GPU resource-limited settings (i.e., RAM, cores, power).

Additionally, we introduced a new training paradigm where we show that using two different optimizers at different phases of training can significantly improve model performance. This result was observed across different architectures, including transformers, ResNets, and token-mixers. DualOpT training needs to be investigated more using different initialisations, hyper-parameter tuning, and learning rates to shed more light into the differences in loss surfaces near and far away from the solution. Additionally, further study on which optimizers are better for initial and final phases also needs to be conducted.  

Among the token-mixing architectures, we found that FNet and MLP-Mixer were not suitable for vision applications but WaveMix and ConvMixer perform at par with the transformer models while consuming less GPU RAM. Wavelets use 2D wavelet transform and ConvMixer uses depth-wise and spatial convolutions, providing inductive priors useful for processing images. This shows that if we use model architectures that can use image priors, it will improve the accuracy, lead to faster training with less data, and consume orders of magnitude less GPU.   

Undoubtedly, further experiments with hyper-parameter tuning, warm up, different data-augmentations, larger datasets and different vision tasks can provide new directions for improving the accuracy of these models in low data, smaller GPU regime. Our results confirm the utility of many of the paths taken towards developing alternatives to purely convolutional or purely attention-based architectures for vision. The success of vision transformers with its impressive results in terms of accuracy comes with orders of magnitude higher costs in terms of training data and GPU RAM, essentially denying the possibility of training such models from scratch using resources available in most settings. The development of hybrid architectures and token-mixers can open up a path to develop alternative model architectures that can provide state-of-the-art results by exploiting the domain-specific inductive priors unique to images (and other specialized data, such as audio) with considerable savings in training data and GPU requirements.


\bibliographystyle{unsrtnat}
\bibliography{references}  

\end{document}